\documentclass[conference]{IEEEtran}
\IEEEoverridecommandlockouts
\usepackage{cite}
\usepackage{amsmath,amssymb,amsfonts}
\usepackage{algorithmic}
\usepackage{graphicx}
\usepackage{textcomp}
\usepackage{xcolor}
\def\BibTeX{{\rm B\kern-.05em{\sc i\kern-.025em b}\kern-.08em
    T\kern-.1667em\lower.7ex\hbox{E}\kern-.125emX}}
\begin{document}

\title{WaterNeRF: Neural Radiance Fields for Underwater Scenes\\
}

\author{\IEEEauthorblockN{Advaith Venkatramanan Sethuraman}
\IEEEauthorblockA{\textit{Robotics Department} \\
\textit{University of Michigan}\\
Ann Arbor, Michigan \\
advaiths@umich.edu}
\and 
\IEEEauthorblockN{Manikandasriram Srinivasan Ramanagopal}
\IEEEauthorblockA{\textit{Robotics Department} \\
\textit{University of Michigan}\\
Ann Arbor, Michigan \\
srmani@umich.edu}
\and
\IEEEauthorblockN{Katherine A. Skinner}
\IEEEauthorblockA{\textit{Robotics Department} \\
\textit{University of Michigan}\\
Ann Arbor, Michigan \\
kskin@umich.edu}
}

\maketitle

\begin{abstract}
Underwater imaging is a critical task performed by marine robots for a wide range of applications including aquaculture, marine infrastructure inspection, and environmental monitoring. However, water column effects, such as attenuation and backscattering, drastically change the color and quality of imagery captured underwater. Due to varying water conditions and range-dependency of these effects, restoring underwater imagery is a challenging problem. This impacts downstream perception tasks including depth estimation and 3D reconstruction. In this paper, we leverage state-of-the-art neural radiance fields (NeRFs) to enable physics-informed novel view synthesis with image restoration and dense depth estimation for underwater scenes. Our proposed method, WaterNeRF, estimates parameters of a physics-based model for underwater image formation and uses these parameters for novel view synthesis. After learning the scene structure and radiance field, we can produce novel views of degraded as well as corrected underwater images. We evaluate the proposed method qualitatively and quantitatively on a real underwater dataset.
\end{abstract}


\section{INTRODUCTION}
Improving underwater perception will advance autonomous capabilities of underwater vehicles, enabling increased complexity of their missions. This will have impacts across marine science and industry~\cite{johnson2017high,saberioon2017application}. However, unlike in air, underwater images are often severely degraded due to water column effects on underwater light propagation. 
In underwater environments, absorption contributes to wavelength-dependent attenuation, resulting in color distortion in the image. Backscattering causes a haze effect across the scene, similar to fog in air~\cite{hazeline}. This can pose a problem for marine robotic systems that rely on accurate and consistent color and dense structural information to enable high level perceptual tasks. Ideally, underwater images could be restored to appear as if taken in air. However, water column effects depend on many factors including scene structure, water characteristics, and light properties. Furthermore, there is often a lack of ground truth for color and structure of underwater scenes. These factors make underwater image restoration and dense scene reconstruction an ill-posed problem. 




On land, there have been impressive advances in neural scene representation, which can learn a volumetric scene function to enable rendering of novel viewpoints. Mildenhall et al. developed Neural Radiance Fields (NeRF), which can be trained on a set of sparse input images to learn a function that maps position and viewing direction to view-dependent radiance and volume density~\cite{mildenhall2020nerf}. 
  These advances in neural scene representation demonstrate the ability to learn scene properties based on a set of input images by learning to model the physics of image formation. We propose to extend these advances to underwater environments by accounting for the effects of underwater light propagation and simultaneously learning a dense structural model of the scene.

\begin{figure*}[t]
    \centering
    \includegraphics[width = 6.3in]{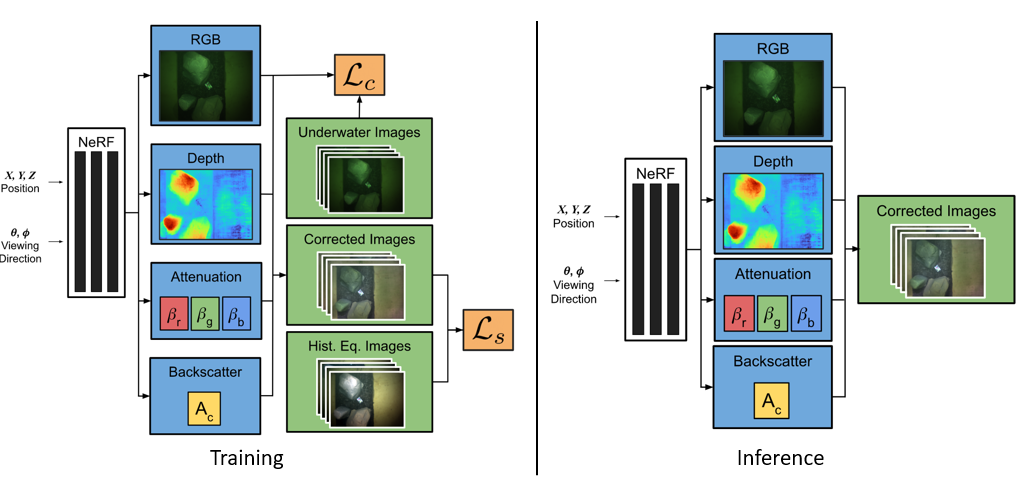}
    \caption{Overview of the WaterNeRF framework. The input for training is a collection of posed images captured in underwater environments.  A multi-layer perceptron learns to map this input to volume density, underwater RGB color, and water column parameters. With the learned volume density, the depth is computed using ray marching. The depth, RGB, and water column parameters are then leveraged to output a corrected RGB image. Loss is computed against the known raw underwater images and a set of reference images obtained by histogram equalization. During inference, the framework can compute an estimate of scene structure and water column parameters to produce corrected color images that have consistent color from varying viewpoints.}
    \label{fig:overview}
\end{figure*}

Our proposed method, WaterNeRF, is a method that leverages neural radiance fields to represent underwater environments. The WaterNeRF framework, shown in Fig.~\ref{fig:overview}, takes as input a corresponding set of monocular underwater and color corrected images, constructs a volume represented as a NeRF, then infers the wavelength-specific attenuation parameters required to restore the scene to its in-air appearance. Notably, WaterNeRF exploits the strengths of NeRFs to encode a scene and uses the resulting depth information to inform a physics-based, data-driven restoration approach.  
\\

Our main contributions are as follows: 
\begin{itemize}
\item Formulation of the underwater color correction problem that leverages implicit scene representations learned through neural radiance fields. 
\item Demonstration of novel view synthesis and 3D reconstruction capabilities of our proposed method. 
\item Evaluation of our method against prior work on a dataset collected in a controlled lab environment with ground truth color. 
\end{itemize}

The remainder of this paper is organized as follows: we present an overview of related works, WaterNeRF's technical approach, and finally our experiments and results. 

\section{Related Work}
\label{sec:related}


\subsection{Image Processing for Underwater Imagery}
Image processing approaches such as histogram equalization~\cite{pizer1987adaptive} and the gray world assumption~\cite{buchsbaum1980spatial} can be used for underwater image enhancement. These approaches act to stretch the effective contrast of underwater images, but are not aware of the underlying physics-based process of underwater image formation. As a result, they do not correct for various range-dependent effects. This can lead to inconsistent image restoration across images of the same scene taken from different viewpoints. 

\subsection{Physics-based Models for Underwater Image Formation}
The attenuation of light transmitted from an object underwater is dependent on water conditions, scene structure, and material properties. Many physics-based image restoration methods first use priors such as the Dark Channel Prior~\cite{carlevaris2010initial}, or haze line prior \cite{hazeline} to estimate the backscatter in an image. For example, Berman \textit{et al.} \cite{hazeline_paper} first estimates backscatter from a single image then produces a transmission map \cite{schechner}. While this method produces appealing results, it requires both a stereo camera setup and knowledge of specific water types. Akkaynak et al. developed an updated model for underwater image formation and restoration~\cite{akkaynak2018revised,akkaynak2017space, akkaynak2019sea}. However, this method requires a known, dense depth map of a scene. 

\subsection{Deep Learning for Underwater Image Restoration}
Recent work has focused on leveraging deep learning to learn a model for underwater image formation to demonstrate underwater image rendering~\cite{ye2022underwater,ugan, li2017watergan}. 
UGAN uses generated data to train a GAN model to remove the degradation present in low quality underwater images~\cite{ugan}. WaterGAN is another adversarial network that considers the inverse problem: generation of synthetic underwater imagery from in-air images~\cite{li2017watergan}. Our work will build on advances in learning-based image restoration, but instead structure our solution to leverage a NeRF framework to enable learning a dense structural model of the scene in addition to corrected color.

\subsection{Neural Radiance Fields}
Most relevant to this work, neural reflectance fields have been developed to enable novel view rendering with scene relighting~\cite{bi2020neural,srinivasan2021nerv}. The effects of light transport through different mediums have also been considered in the context of NeRFs. Specifically, NeRFs have been used to estimate albedo and ambient light from satellite imagery \cite{satnerf}. Recently, NeRFs have been used to model the refraction of light at the interface between air and water \cite{Wang2022NeReFNR}. Similarly, a NeRF formulation for our problem is convenient because it provides a framework for optimization of attenuation parameters as a function of corrected color and ray termination distances. 

\section{Technical Approach}
\label{sec:tech}
Figure~\ref{fig:overview} shows an overview of the proposed pipeline for the WaterNeRF framework. During training, the network takes as input 3D position and viewing direction of a scene point and learns to output volume density, underwater RGB color, and water column parameters, including attenuation and backscatter parameters. The depth, underwater RGB, and water column parameters are input to a physics-based model for underwater image restoration that outputs a corrected RGB image. The training images require known poses, which can be obtained by a Structure from Motion (SfM) pipeline such as COLMAP~\cite{fisher2021colmap}. During inference, we are able to render images from novel viewpoints with and without water column effects.

\subsection{WaterNeRF Formulation}
NeRFs take as input a query location $x, y, z$ and viewing directions $\theta, \phi$ and output the color $r,g,b$ and density $\sigma$ \cite{mildenhall2020nerf}. We specifically leverage mip-NeRF, which replaces rays with conical frustums to avoid aliasing issues~\cite{barron2021mipnerf}. Mip-NeRF was chosen because both image quality and depth map quality are crucial to image restoration underwater. 

Mip-NeRF uses a multilayer perceptron (MLP) parameterized by weights $\Theta$. A ray is defined as $r(t) = o + td$, where $t \in [t_n, t_f]$ lies between near plane $t_n$ and far plane $t_f$, $o$ is the ray origin, and $d$ is the ray direction. The query location is encoded using an integrated positional encoding $\gamma(r(t))$ before being passed to the MLP \cite{barron2021mipnerf}. 
\begin{equation}
    \forall t_k \in t, [\sigma_k, c_k] = MLP(\gamma(r(t_k)), \theta, \phi; \Theta)
\end{equation}
Each $\sigma_k, c_k$ is used in a numerical quadrature sum to estimate the color $C(r; \Theta, t)$ of a ray. 
\begin{equation}
    C(r; \Theta, t) = \sum_k T_k(1-exp(-\sigma_k(t_{k+1}-t_{k})))c_k
\end{equation}
The depth at a given ray can be found similarly: 
\begin{equation}
    D(r; \Theta, t) = \sum_k T_k(1-exp(-\sigma_k(t_{k+1}-t_{k})))t_{km}
\end{equation}
\noindent where 
    $t_{km} = \frac{1}{2}(t_{k+1} + t_{k})$ and

\begin{equation}
    T_k = exp \left(-\sum_{k'<k}\sigma_{k'}(t_{k'+1} - t_{k'})\right)
\end{equation}

Finally, a color loss $\mathcal{L}_c$ is computed between the predicted color $C(r;\Theta, t)$ and the ground truth pixel values $C^*(r)$ from the observed images, which in our case are the original underwater images. This loss is composed of coarse $C(r;\Theta, t^c)$ and fine $C(r;\Theta, t^f)$ queries to a single MLP network. 
\begin{equation}
    \mathcal{L}_c = \lambda||C^*(r)-C(r;\Theta, t^c)||^2_2 + ||C^*(r)-C(r;\Theta, t^f)||^2_2
    \label{mse_nerf}
\end{equation}

\subsection{Color Correction Branch}
In order to determine the color corrected appearance of the scene, we use the model for light transport in water presented in \cite{schechner}. $I_c(x)$ is the acquired image at pixel location $x$ for channel $c \in \{R, G, B\}$, $t_c(x)$ is the transmission map, $J_c(x)$ is the object's original radiance prior to attenuation, and $A_c$ is the veiling light or backscatter. The transmission map $t_c(x)$ is related to the wavelength specific attenuation coefficients $\beta_c$ and the depth at the pixel $D(x)$.
\begin{equation}
    I_c(x) = t_c(x)J_c(x) + (1-t_c(x))A_c 
    \label{trans}
\end{equation}
\begin{equation} 
t_c(x) = exp(-\beta_cD(x))
\end{equation}

Using this relationship, we can define the color corrected output of the network $J(r; \Theta, t)$ as shown in Fig. \ref{fig:overview}. 
\begin{equation}
    J(r; \Theta, t) = \frac{C(r; \Theta, t) - (1-t_c(x))A_c}{t_c(x)}
    \label{relation}
\end{equation}

Since we do not have ground truth color corrected images $I^{\ast}(r)$, we cannot directly minimize a loss between $J(r; \Theta, t)$ such as in Equation (\ref{mse_nerf}). Instead, we can rely on a supporting dataset of histogram equalized underwater images $H(r)$ to provide an initialization. However, histogram equalized imagery does not respect Equation (\ref{trans}) nor does it produce consistently corrected images across the scene. To address these problems, we reframe the color correction loss as an optimal transport problem and attempt to match the distribution of pixels in histogram equalized images. We wish to estimate the attenuation coefficients $\beta_c$ and veiling light $A_c$ to correct an underwater image. 

\subsection{Optimal Transport Primer}
Discrete optimal transport problems involve distributions such as $\mu, \nu \in \mathcal{P}(X)$ where $\mu = \sum_{i=1}^{n}a_i\delta_{x_i}$ and $\nu = \sum_{j=1}^{m}b_j\delta_{y_j}$ and $\delta$ is a Dirac-delta centered at points $\{x_i | \forall i \in [1,n]\}$ and $\{y_j | \forall j \in [1,m]\}$ \cite{ot_background}. The weights associated with this discrete distribution are such that $a = (a_1, \dots, a_n)^\top \in \Delta_n$ and $b = (b_1, \dots, b_m)^\top \in \Delta_m$, where 
\begin{equation}
    \Delta_n = \{p \in \mathbb{R}^n_{+} \ | \ p^\top\textbf{1}_n = 1 \}
\end{equation}
$\textbf{1}_n$ is a vector containing ones of length $n$. The goal is to compute a transport plan $T$ such that the Wasserstein distance of the following form is minimized: 
\begin{equation}
    W^p_p(\mu, \nu) = \min_{T\in\Pi(a,b)}\langle T, M \rangle
\end{equation}

$M$ is a cost matrix $\mathbb{R}^{n\times m}$ where $M_{ij} = d(x_i, y_j)$, a valid distance between points $x_i$ and $y_j$. $\langle T, M \rangle$ is the Frobenius product between matrices $Tr(T^\top M)$. The transport plan $T \in\Pi(a,b)$ must be within the set: 
\begin{equation}
\Pi(a, b) = \{T\in\mathbb{R}^{n\times m}_{+}: T\mathbf{1}_m = a, T^\top\mathbf{1}_n = b\}
\end{equation}

This expensive optimization problem can be regularized and solved using a differentiable process called the Sinkhorn algorithm $\tilde{S}_\lambda$. The Sinkhorn algorithm is considered an efficient approximation of the Wasserstein distance~\cite{sinkhorn1967diagonal,cuturi2013sinkhorn}. 
\begin{equation}
    \tilde{S}_\lambda = \min_{T\in\Pi(a,b)}\langle T,M \rangle - \frac{1}{\lambda}h(T) 
\end{equation}
\begin{equation}
    h(T) = -\sum_{i,j=1}^{n,m}T_{i,j}(logT_{ij}-1)
\end{equation}

In this work, the Sinkhorn Loss is used between probability distributions $\mu$, which denotes the distribution of corrected colors output by WaterNeRF, $J(r; \Theta, t)$, and $\nu$, which denotes the distribution of colors in the histogram equalized images. Both $\mu$ and $\nu$ are distributions in the RGB-space $\mathbb{R}^3$. Formally, the optimization problem becomes: 

\begin{equation}
\begin{aligned}
\beta_c^*, A^*_c &=  \arg\min_{\beta_c, A_c}\tilde{S}_\lambda(\mu, \nu)\\
\textrm{such that } \mu &= \frac{\nu - (1-t_c(x))A_c}{t_c(x)}
\end{aligned}
\end{equation}


The framework must optimize parameters $\beta_c$ in order to minimize the Sinkhorn Loss between two distributions of colors. Since the optimization is physically constrained, the optimal point $\beta_c^*$ will provide the attenuation parameters required to map in-water images to histogram equalized images while respecting the geometric (depth map) and physical (attenuation) properties of the image formation model in Equation (\ref{trans}). This formulation allows the model to infer parameters $\beta^*_c, A^*_c$ that yield a color corrected image. 
The Sinkhorn loss used in this work is the following:
\begin{equation}
    \mathcal{L}_s = \tilde{S}_\lambda(J(r; \Theta, t), H(r))
\end{equation}
The final objective is: 
\begin{equation}
    \mathcal{L} = \mathcal{L}_c + \alpha \mathcal{L}_s
\end{equation}

\vspace{7mm}

\begin{table*}[h]
\begin{center}
\caption{Angular error (degrees) in color correction for novel views (N=3). Lower is better ($\downarrow$). Best is shown in \textbf{bold} \label{color_corr}}
\begin{tabular}{lcllllll}
\textbf{Method}        & \multicolumn{1}{l}{\textbf{Blue $\bar{\psi}$}} & \textbf{Red $\bar{\psi}$}              & \textbf{Magenta $\bar{\psi}$}          & \textbf{Green $\bar{\psi}$}            & \textbf{Cyan $\bar{\psi}$}             & \textbf{Yellow $\bar{\psi}$}           & \textbf{Mean $\bar{\psi}$}                     \\ \hline
Histogram Equalization & 6.67                              & \multicolumn{1}{c}{\textbf{14.07}} & \multicolumn{1}{c}{8.06}  & \multicolumn{1}{c}{\textbf{9.55}}  & \multicolumn{1}{c}{3.78}  & \multicolumn{1}{c}{\textbf{3.11}}  & \multicolumn{1}{c}{\textbf{7.54}} \\
Underwater Haze Lines  & \textbf{4.39}                              & \multicolumn{1}{c}{16.09} & \multicolumn{1}{c}{\textbf{5.17}}  & \multicolumn{1}{c}{13.91} & \multicolumn{1}{c}{8.80}  & \multicolumn{1}{c}{4.68}  & \multicolumn{1}{c}{8.84}          \\
FUnIE-GAN              & 36.39                             & \multicolumn{1}{c}{16.83} & \multicolumn{1}{c}{25.96} & \multicolumn{1}{c}{10.08} & \multicolumn{1}{c}{23.23} & \multicolumn{1}{c}{12.04} & \multicolumn{1}{c}{20.75}         \\
HENeRF     & 5.12                             & \multicolumn{1}{c}{14.89} & \multicolumn{1}{c}{7.72}  & \multicolumn{1}{c}{11.13} & \multicolumn{1}{c}{4.08}  & \multicolumn{1}{c}{5.59}  & \multicolumn{1}{c}{7.78}          \\
WaterNerf (ours)       & 7.88                              & \multicolumn{1}{c}{19.51} & \multicolumn{1}{c}{9.71}  & \multicolumn{1}{c}{11.32} & \multicolumn{1}{c}{\textbf{2.80}}  & \multicolumn{1}{c}{5.59}  & \multicolumn{1}{c}{9.47}          \\
\end{tabular}
\end{center}
\end{table*}
\section{Experiments \& Results}
\label{sec:exp}
\subsection{Training Details}
The network was trained in JAX, modified from the mip-NeRF repository \cite{barron2021mipnerf}\cite{jax2018github}. 
An initial learning rate of 5e-4 and a final learning rate of 5e-6 are used with learning rate decay. The Adam optimizer is used \cite{adam}. The networks are trained for 600,000 iterations, which takes around 8 hours on an Nvidia GeForce RTX 3090 GPU. We use a $\alpha = 0.5$ for weighting the Sinkhorn Loss. 

\subsection{Dataset}
We evaluated our system on the open source UWBundle dataset, which consists of 36 images of an artificial rock platform submerged in a controlled in-lab water tank environment \cite{Skinner:2016ab}. Multiple views of the scene were captured with an underwater camera in a lawnmower pattern, common in underwater surveying. The rock platform contains a colorboard in the scene, making it ideal for performing quantitative evaluation of color correction. An in-air ground truth 3D reconstruction was made using KinectFusion \cite{izadi2011kinectfusion}. 


\subsection{Comparison to Baselines}
We compare the performance of WaterNeRF to prior work on image processing, physics-based underwater image restoration,  and deep learning-based methods.

\textbf{FUnIE-GAN --} This is a conditional GAN-based model that learns the mapping between degraded underwater imagery and restored images \cite{funie}.

\textbf{Underwater Haze Lines Model --} Using the haze lines prior, this model estimates backscatter light and transmission map. Then, attenuation coefficients are chosen from Jerlov's classification of water types based on performance under the Gray-World assumption \cite{jerlov, hazeline_paper}. 

\textbf{Histogram Equalization --} We use histogram equalized imagery to provide an initialization point for our optimization algorithm and we compare to these images as a baseline. Histogram equalization is performed separately for each color channel.

\textbf{HENeRF --} This is a mipNeRF trained directly on the histogram equalized images. This has no color correction module and uses an MSE loss on the histogram equalized pixel values. 

\textbf{WaterNeRF-No Sinkhorn --} This is an ablation on the Sinkhorn loss. The WaterNeRF is trained with an MSE loss between the corrected color prediction $J(r; \Theta, t)$ and the histogram equalized color $H(r)$.


\begin{figure*}
\centering
\includegraphics[width = 6.0in]{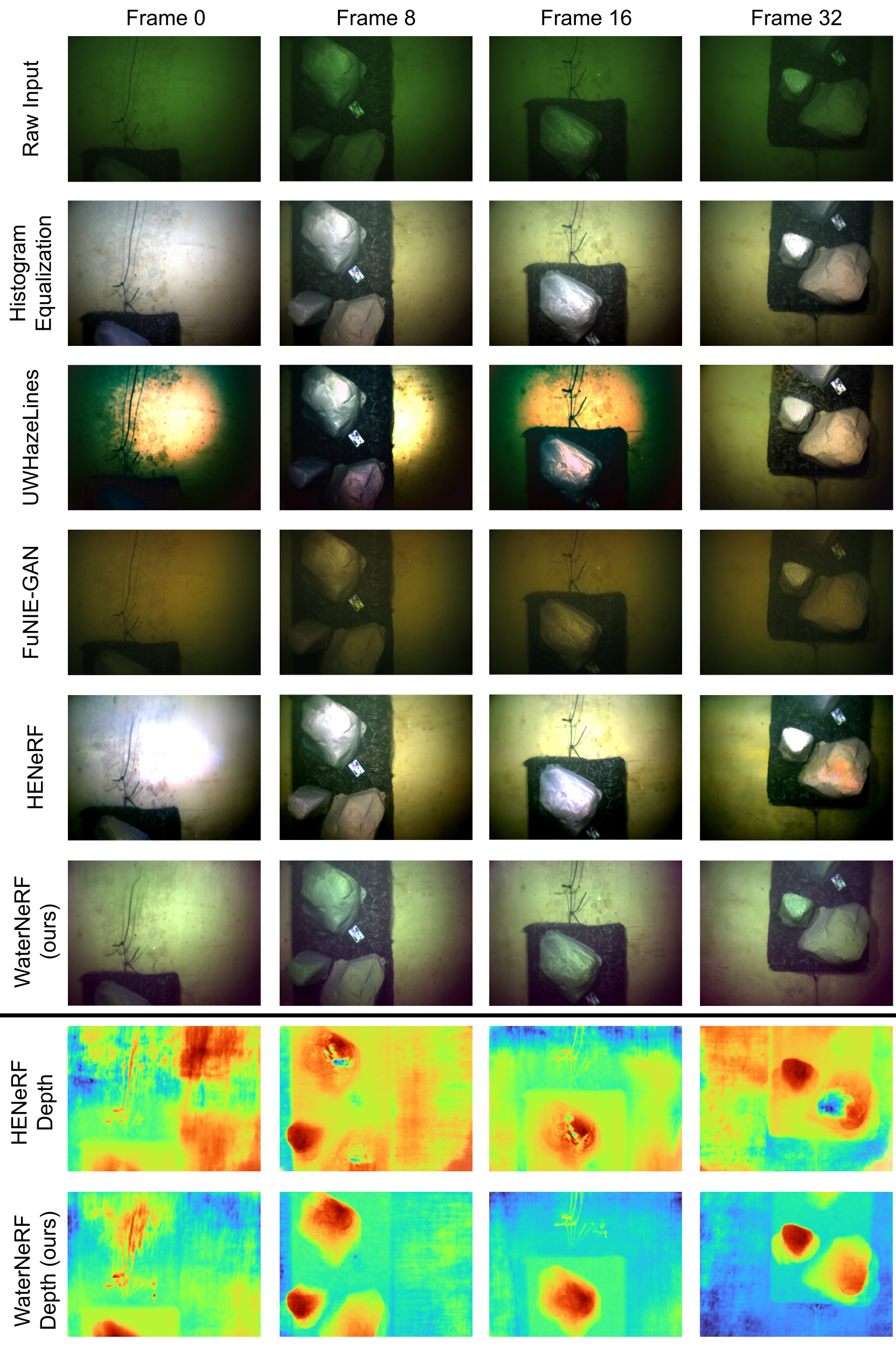}
\caption{(top) Raw images, (second row) histogram equalized images, (third row) UWHazeLines output, (fourth row) FuNIE-GAN output, (fifth row) HENeRF Output, (sixth row) WaterNeRF Output, (seventh row) Depth Output from HENeRF, (eight row) Depth Output from WaterNeRF  \vspace{30mm}}
\label{fig:qual_mhl}
\end{figure*}

\subsection{Qualitative Evaluation} 
Figure~\ref{fig:qual_mhl} shows qualitative performance against the baselines on the UWBundle dataset. Histogram equalization and HENeRF result in visually appealing images, but the corrected images are not consistent across multiple viewpoints. The Haze Lines model suffers from vignetting artifacts. WaterNeRF and FUnIE-GAN both produce consistent color corrected images across multiple images, however FUnIE-GAN does not perform depth estimation. HENeRF and WaterNerF both produce dense maps of the scene. Since the histogram equalized images are inconsistent, HENeRF produces poor depth structure with holes. WaterNeRF produces dense depth maps without visible discontinuities. 

\subsection{Color Correction Evaluation} 
Our primary evaluation metric for color correction is average angular error $\bar{\psi}$ in the pixel space \cite{hazeline_paper}. $I(x_i)$ is the color at pixel location $x_i$ and $\langle R_g,G_g,B_g \rangle$ is the color at the ground truth pixel location. The colors are normalized then the angle in degrees between the two unit vectors is computed by:  
\begin{equation}
     \bar{\psi} = \textrm{cos}^{-1}\left(\frac{I(x_i)\cdot \langle R_g,G_g,B_g \rangle}{\|{I(x_i)}\| \cdot \|\langle R_g,G_g,B_g \rangle\|} \right)
\end{equation}

Table~\ref{color_corr} shows the average angular error for patches of the colorboard within corrected images from three held out test images containing the colorboard. The error was calculated between patches of the colorboard imaged in air and patches within corrected images underwater. For each color, a rectangular region was taken from the patch and averaged. This average color was used to calculate $\bar{\psi}$. Histogram equalization performs the best for color accuracy. However, it is important to note that the histogram equalization process will be biased for images that contain the color board. FUnIE-GAN suffers on color accuracy relative to other methods. WaterNeRF is able to correct images with comparable accuracy to histogram equalization and the Haze Lines model.




\subsection{Underwater Image Quality}
Next, we evaluate the quality of restored images using the Underwater Image Quality Metric (UIQM) \cite{uiqm}. UIQM considers the Underwater Image Colorfulness Measure (UICM), the Underwater Image Contrast  Measure (UIConM), and the Underwater Image Sharpness Measure (UISM): 
\begin{equation}
UIQM = c_1 \times UICM + c_2 \times UISM + c_3 \times UIConM
\end{equation}

We use the coefficients: $c_1 = 0.282, c_2 = 0.2953, c_3 = 3.5753$ as in \cite{uiqm}. Table~\ref{uiqm} reports the UIQM for corrected images produced by the baselines and WaterNeRF on the UWBundle dataset. There are 5 held out test images, 3 containing the colorboard and 2 with no colorboard. WaterNeRF produces images of higher underwater image quality than the Haze Lines model and FUnIE-GAN, but lower quality than histogram equalization.  

\begin{table}[h]
\begin{center}
\caption{UIQM performance on novel views within the UWBundle dataset (N=5). Higher is better ($\uparrow$). Best is shown in \textbf{bold}. \label{uiqm}}
\begin{tabular}{lc}
\textbf{Method}        & \multicolumn{1}{l}{\textbf{UIQM}} \\ \hline
Histogram Equalization &  \textbf{2.96}                             \\
Underwater Haze Lines  &  2.60                             \\
FUnIE-GAN              &  2.06                             \\
HENeRF & 2.40 \\ 
WaterNeRF (ours)       &  2.88                            
\end{tabular}
\end{center}
\end{table}

\subsection{Scene Consistency Evaluation}
In order to quantify the consistency of the color corrections, we calculate the average standard deviation of the intensity-normalized RGB values of pixels tracked through the scene. We call this the scene consistency metric (SCM). First, we find a set of SURF features $\mathcal{P}$ between frames in the underwater images, where $N = |\mathcal{P}|$ \cite{surf}. Then we track the pixels $x_i \in \mathbb{R}^3, i \in [1, N_x]$ corresponding to a feature $x \in \mathcal{P}$ through $N_x$ corrected images. Finally, we calculate the standard deviation of the RGB values of those pixels. This metric provides a way to quantify the consistency of the correction method across different views within the same scene.
\begin{equation} 
SCM = \frac{1}{N}\sum_{x \in \mathcal{P}}\sqrt{\frac{\sum_{x_i \in x}(x_i - \mu_x)^2}{N_x}}
\end{equation}

Table~\ref{scm} reports the Scene Consistency Metric (SCM). Each color channel is reported individually. FUnIE-GAN performs well on the SCM metric.  WaterNeRF also produces consistent images across scenes, with the lowest SCM in the red channel and second lowest in the blue channel. 

\begin{table}[h]
\begin{center}
\caption{Scene Consistency Metric (SCM) across novel views in the UWBundle dataset (N=5). Lower is better ($\downarrow$). Best is shown in \textbf{bold}. \label{scm}}
\begin{tabular}{lcll}
\textbf{Method}        & \multicolumn{1}{l}{\boldmath{$SCM_R$}} & \boldmath{$SCM_G$} & \boldmath{$SCM_B$} \\ \hline
Histogram Equalization & 0.1263                           & 0.1287       & 0.1120       \\
Underwater Haze Lines  & 0.3840                           & 0.3426       & 0.3068       \\
FUnIE-GAN              & 0.1064           & \textbf{0.0619}       & \textbf{0.0908}       \\
HENeRF & 0.2051 & 0.1731 & 0.1800 \\
WaterNeRF (ours)       & \textbf{0.0423}       & 0.1975       & 0.0919      
\end{tabular}
\end{center}

\end{table}

\subsection{Evaluation of Depth Maps}
\begin{figure}[t]
\begin{center}
\begin{tabular}{cc}
{\includegraphics[width = 1.7in]{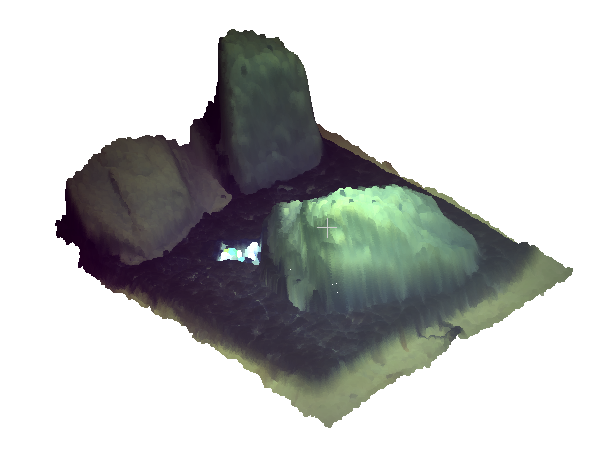}} &{\includegraphics[width = 1.5in]{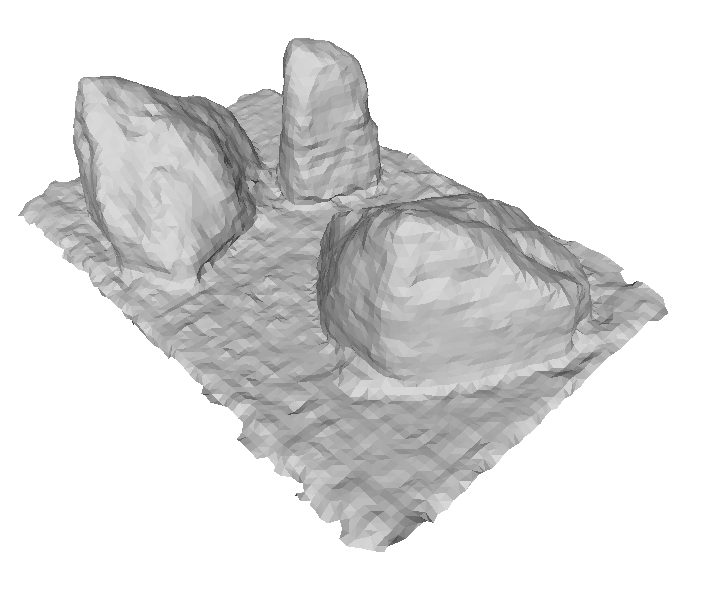}}

\end{tabular}
\end{center}
\caption{(left) 3D reconstruction from WaterNeRF. Color corrected RGB points are projected onto the pointcloud provided from the dense depth map. (right) Ground truth in-air 3D scan. \label{pointclouds}}
\label{fig:qual_bios}
\end{figure}

In addition to underwater and corrected color images, WaterNeRF also produces a dense depth map for each novel view generated. To evaluate the quality of this reconstruction, we project the 2D pixels into 3D points using the camera's intrinsic matrix.  Finally, we use CloudCompare~\cite{girardeau2015cloud} to compute the distance between the point cloud and a ground truth mesh for the UWBundle dataset. Table~\ref{dist} reports the evaluation between a point cloud of the scene generated by WaterNeRF and a ground truth mesh. WaterNeRF demonstrates higher accuracy on depth reconstruction than HENeRF. Figure \ref{pointclouds} shows sample point clouds projected from the dense depth estimates and color corrected imagery for WaterNerf. 

\begin{table}[h]
\begin{center}
\caption{RMSE in meters between a ground truth 3D mesh for the UWBundle dataset and the pointclouds generated by WaterNeRF and HENeRF.\label{dist}}
\begin{tabular}{lcc}
Model     & $Pred \rightarrow GT$ & G$T \rightarrow Pred$ \\ \hline
WaterNeRF & \textbf{0.036 }     & \textbf{0.0014}      \\
HENeRF    & 0.105      & 0.099      
\end{tabular}

\end{center}

\end{table}




\subsection{Ablation: WaterNeRF-No Sinkhorn Loss}
\begin{figure}[h]
    \centering
    \includegraphics[width=3in]{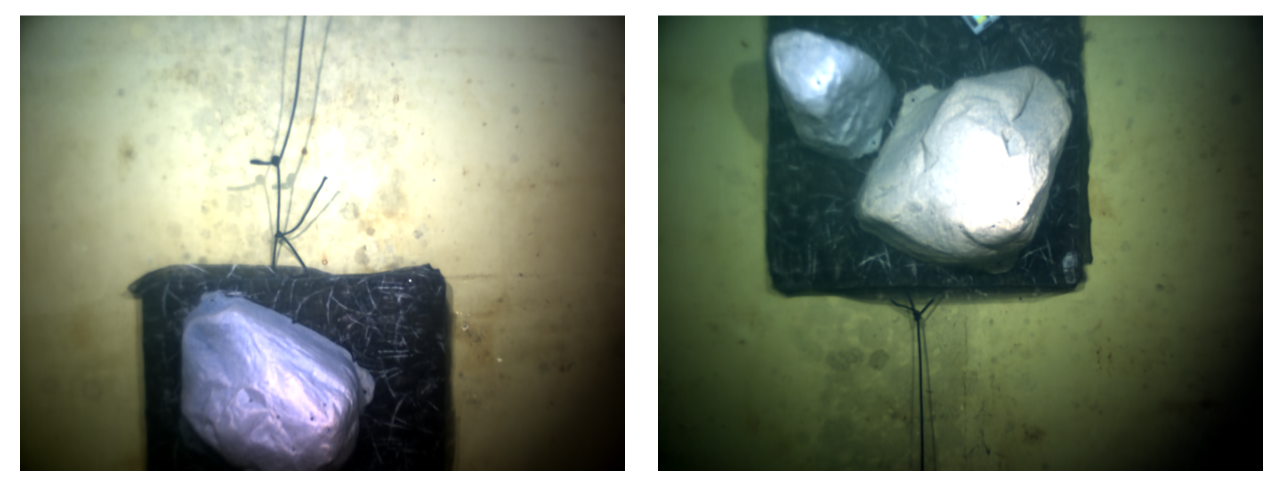}
    \caption{Two views generated by WaterNeRF with no Sinkhorn loss results in inconsistent restorations.}
    \label{fig:ns}
\end{figure}

We performed ablation studies with no Sinkhorn loss to demonstrate that a direct MSE loss on the histogram equalized images can produce inconsistent views (Fig.~\ref{fig:ns}).


\section{Conclusion \& Future Work}
\label{sec:concl}
This work is the first to leverage neural volume rendering for underwater scenes to enable novel view synthesis of underwater scenes with full scene restoration and dense depth estimation. Notably, WaterNeRF's improved scene consistency and 3D reconstruction ability demonstrate the importance of learning attenuation parameters for novel view synthesis. 
Future work includes using 3D object information and color from WaterNeRF to inform underwater manipulation and exploration tasks for marine robotic platforms. 

\bibliographystyle{ieeetr}
\bibliography{ref}

\end{document}